\documentclass[letterpaper, 10 pt, conference]{IEEEtran}  

\IEEEoverridecommandlockouts                              

\usepackage{cite}
\usepackage{amsmath,amssymb,amsfonts}
\usepackage{algorithmic}
\usepackage{graphicx}
\usepackage{textcomp}
\usepackage{xcolor}
\usepackage{xurl}
\usepackage{booktabs}
\usepackage{float}
\usepackage{adjustbox}
\usepackage{pythonhighlight}
\usepackage{listings}
\usepackage{fancyhdr}
\usepackage{array}
\usepackage[ruled,vlined]{algorithm2e}
\usepackage{stfloats}
\usepackage[most]{tcolorbox}
\usepackage{orcidlink}
\usepackage{multirow}

\begin{document}

\title{\LARGE \bf
LLM-Driven Corrective Robot Operation Code Generation with Static Text-Based Simulation}


\author{Wenhao Wang$^{1}$, Yi Rong$^{1}$, Yanyan Li$^{2}$, Long Jiao$^{1}$, Jiawei Yuan$^{1}$
\thanks{$^{1}$Department of CIS, University of Massachusetts Dartmouth
        {\tt\small \{wwang5, yrong, ljiao, jyuan\}@umassd.edu}}%
\thanks{$^{2}$Department of CSIS, California State University San Marcos.
        {\tt\small yali@csusm.edu}}%
\thanks{This work is supported by the US National Science Foundation awards 2318710 and 2318711.}
\thanks{Project is available at \url{https://github.com/ai-uavsec/LLM-Driven-Static-Simulation}}
}

\maketitle \thispagestyle{fancy}

\begin{abstract}
Recent advances in Large language models (LLMs) have demonstrated their promising capabilities of generating robot operation code to enable LLM-driven robots. To enhance the reliability of operation code generated by LLMs, corrective designs with feedback from the observation of executing code have been increasingly adopted in existing research. However, the code execution in these designs relies on either a physical experiment or a customized simulation environment, which limits their deployment due to the high configuration effort of the environment and the potential long execution time. In this paper, we explore the possibility of directly leveraging LLM to enable static simulation of robot operation code, and then leverage it to design a new reliable LLM-driven corrective robot operation code generation framework. Our framework configures the LLM as a static simulator with enhanced capabilities that reliably simulate robot code execution by interpreting actions, reasoning over state transitions, analyzing execution outcomes, and generating semantic observations that accurately capture trajectory dynamics. To validate the performance of our framework, we performed experiments on various operation tasks for different robots, including UAVs and small ground vehicles. The experiment results not only demonstrated the high accuracy of our static text-based simulation but also the reliable code generation of our LLM-driven corrective framework, which achieves a comparable performance with state-of-the-art research while does not rely on dynamic code execution using physical experiments or simulators.
\end{abstract}

\section{INTRODUCTION}

Robots are being increasingly deployed to execute tasks based on human instructions. However, designing a robot that has intelligence to perform complex tasks reliably remains challenging, as it demands both robust instruction interpretation and executing tasks with robust reasoning. Recent advances in LLMs~\cite{GPT4, Gemini, deepseek} have demonstrated remarkable proficiency in robotic areas such as control~\cite{Real}, planning~\cite{planning}, and navigation~\cite{L3MVN}. Their strong context understanding and generation capabilities enable the robot to comprehend human instructions and generate corresponding robot operation code~\cite{ChatGPTRobotics, TypeFly, CodeasPolicies, GSCE}, thereby greatly simplifying the process of robot programming.


Unlike text generation applications, where semantic-level accuracy is sufficient, executing logically inconsistent or syntactically incorrect code on a robot can lead to unexpected outcomes and even unsafe robot behaviors, such as UAV crashes. To improve the reliability of LLMs' robot operation code generation, recent studies have incorporated a corrective process that leverages the observation of executing LLM-generated code in a physical experiment or simulator to identify issues and perform refinement \cite{AutoTAMP,InteractivePlanning,CLGSCE}. While these corrective designs have shown their effectiveness in enhancing the reliability of LLM-generated operation code, they still face challenges from different aspects. Specifically,  the configuration of a physical or simulation environment for robot code execution requires specialized expertise to support the tasks (e.g., designing scenes and modeling a robot's replica), especially for custom-built robots. For example, the code execution in ref \cite{CLGSCE} requires a specific design of mapping that transforms numerical state representation to the semantic description, which needs to be customized case-by-case. In addition, existing corrective designs require multiple rounds of interactive refinements, which can lead to a long execution time in the experiment or simulation when the task is complicated or involves time-consuming operations (e.g., monitor the area for 15 minutes). Therefore, removing the dependence on dynamic execution in a physical experiment or simulator while still maintaining the corrective feedback and refinement features for LLM-driven robot operation code generation becomes a challenging gap to close.

\begin{figure}[t]
    \centering
    \includegraphics[width=.8\linewidth]{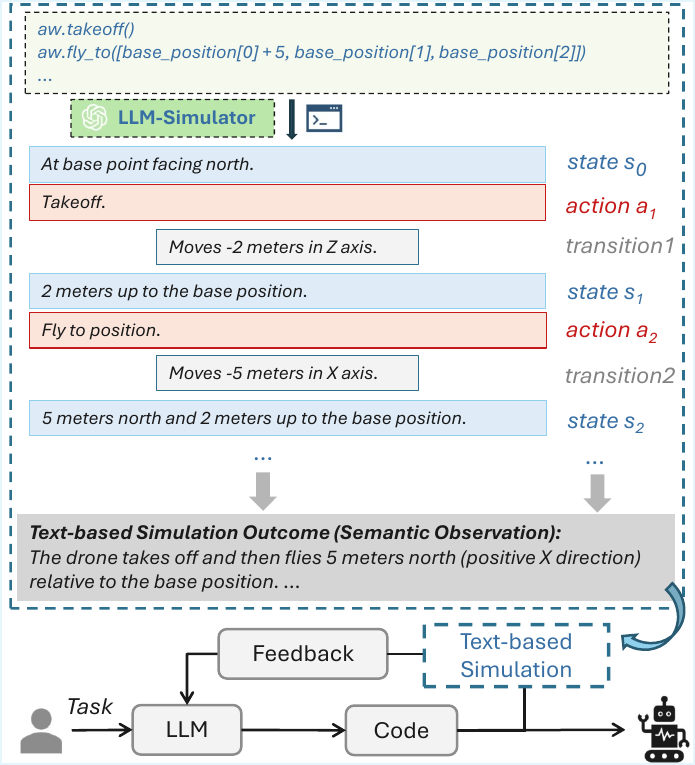}
    \caption{An overview of LLM-driven corrective robot operation code generation with static text-based simulation.}
    \label{fig:overall}
\end{figure}

In this paper, we explore the possibility of directly employing LLMs for static text-based simulation over robot operation code and obtaining effective semantic observation to enable the refinement of LLM-driven robot operation code generation. Specifically, we propose a novel static text-based simulation solution powered by LLM to statically simulate the execution of robot operation code and generate accurate simulation outcomes. As shown in Fig.~\ref{fig:overall}, our static simulation solution emulates the code execution by interpreting the actions encoded in the text of the code, reasoning about the corresponding state and environment-driven transitions, analyzing the next robot state after the action execution, and ultimately producing semantic observations that capture the robot's trajectory dynamics. On top of that, we propose our corrective robot code generation with static text-based simulation that unfolds as follows: 1) An LLM configured as a code generator first generates the initial version of robot operation code based on the task description from the user; 2) The LLM-based simulator ``executes'' the code and produces a semantic observation of the robot's trajectory; 3) An evaluator LLM analyzes the observation and identifies mismatches between the task description and the robot's trajectory, and then provides feedback that depicts the mismatched actions; 4) Guided by the feedback, the code generator corrects the mismatches. This iterative correction process continues until the evaluation confirms alignment between the code and task objectives, after which the final version of code is deployed to the robot for task execution.



We extensively evaluated our proposed framework for the operation code generation of UAV tasks with various levels of task complexities. Our results show that our static text-based simulation achieves over $97.5\%$ of simulation accuracy compared with the widely adopted UAV simulators, i.e., AirSim \cite{AirSim} and PX4-Gazebo \cite{PX4}. In addition, our corrective code generation framework delivers comparable robot execution performance as the state-of-the-art (SOTA) method relying on dynamic code execution in physical experiment or simulator, with an $85\%+$ success rate and $96.9\%+$ completeness on different UAV systems, i.e., $96.9\%+$ of required actions in all evaluated tasks are completed correctly and $85\%+$ of tasks are entirely completed without any error. To demonstrate the adaptability of our framework and its performance on real-world robot deployment, we also evaluated it for the operation code generation for UAVs and ground robots. Our experiment results show that our framework can also achieve high success rates ($87.5\%+$) and completeness ($96.9\%+$).




\section{RELATED WORK}
\subsection{LLM-Driven Corrective Robot Code Generation}
Using LLMs to generate operation codes for robot tasks has become a prevalent trend in recent research. By configuring LLMs with appropriate system prompting techniques, existing research has shown that LLMs have the potential to generate robot operation codes ~\cite{ChatGPTRobotics, TypeFly, CodeasPolicies, GSCE}. To further enhance the reliability of LLMs' output and support more complicated robot tasks, recent efforts have adopted corrective code generation such that the errors or mismatches are iteratively detected and corrected; therefore, the robot could perform the desired task accurately when executing the final code~\cite{CoPAL, CAPE, trust}. However, the physical execution during the correction process may increase the hardware cost and raise safety risks, as executing code with errors could cause irreparable damage to robots, such as UAV crashes. More recent studies proposed a simulation-based pipeline that eliminates the potential risks during physical execution~\cite{CLGSCE, AutoTAMP, hu2024robo}. For example, \cite{CLGSCE} leverages the AirSim~\cite{AirSim} simulator to iteratively correct the UAV operation code until the code is ready for deployment. The authors also develop semantic observations rather than numerical representation~\cite{InteractivePlanning} to describe the UAV trajectory to further improve performance. Robo-Instruct~\cite{hu2024robo} fine-tuning LLM that checks LLM-generated robot programs with a simulator and revises them until correct. 


\subsection{LLM-based Simulation}
Recent advances in LLMs~\cite{GPT4, Gemini, deepseek} have demonstrated their exceptional capabilities in world modeling. Previous studies have explored the integration of LLMs into agent-based modeling and simulation within the social~\cite{LLM-simulator-survey} and planning~\cite{yang2024evaluating} domains. For example, BeSimulator~\cite{besimulator} built an LLM-powered framework towards behavior simulation on the behavior tree. However, studies on text game show that current LLMs are not yet able to reliably act as text world simulators as they are likely to make errors when arithmetic, common-sense, or scientific knowledge is needed~\cite{LLM-simulator}. This limitation highlights the need for further research to examine the use of LLMs for the robot code simulation domain and develop methods to enhance the ability of LLMs to perform accurate and reliable simulations of robot operation code. 

\subsection{Prompt Engineering}
Prompt engineering can effectively communicate and interact with LLM-driven tools~\cite{PromptEngineering}. Recent studies have increasingly used prompt engineering to improve the reliability of LLM-driven robotic systems. For example, PromptBook proposes a prompt framework as a system prompt to enhance the code generation~\cite{PromptBook}. On the one hand, prompt engineering leverages in-context learning~\cite{URIAL} and few-shot learning~\cite{llm-fewshot} to enable LLM to learn knowledge within the given context and identify patterns from a limited set of examples. These techniques facilitate the generation of code that adheres to robot policy and how to ground task descriptions from a few examples~\cite{CodeasPolicies}. On the other hand, CoT~\cite{CoT} prompts the LLM to articulate intermediate inference steps, which is valuable for robotic tasks that require sequential, stepwise decision-making. CoT encourages the production of code that aligns with each stage of the intended action plan. Previous studies have embedded CoT within the examples to guide LLMs through reasoning step-by-step ~\cite{ISR-LLM, PromptBook}. In this study, we utilize prompt engineering strategies to facilitate our text-based simulation.


\begin{figure*}[!htbp]
    \centering{\includegraphics[scale=0.74]{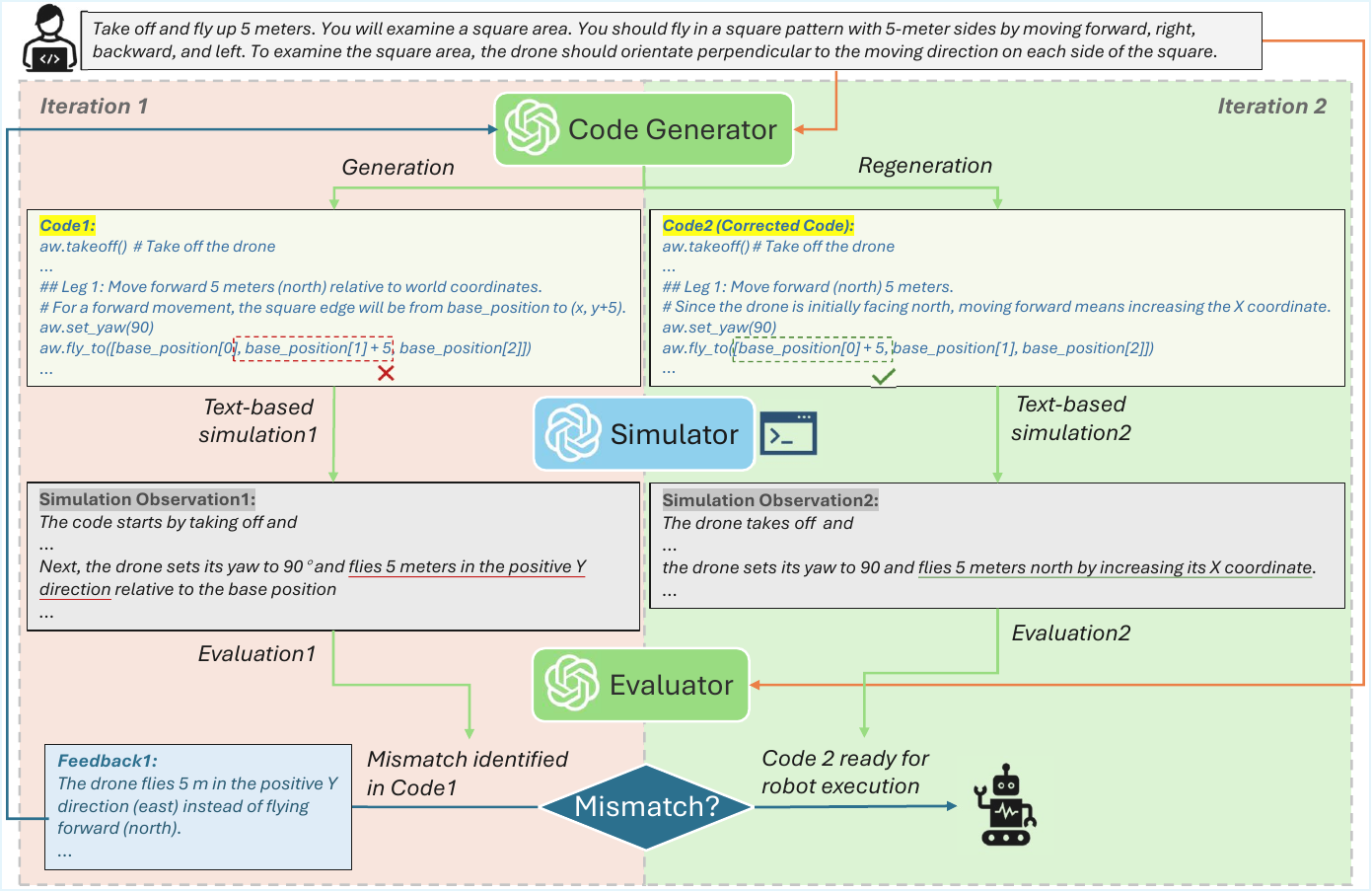}}
    \caption{An illustrative example of corrective code generation with text-based simulation. In the first iteration, the LLM-based simulator accurately produces an observation of UAV actions, while the evaluator identifies the mismatch and constructs feedback. Based on the feedback, the code generator corrects the mismatch and produces a valid code for robot operation in the second iteration.
    }
    \label{fig:design}
\end{figure*}

\section{METHOD} \label{sec:method}

\subsection{Overview} \label{sec:overview}
Fig.~\ref{fig:design} presents the overall framework of our corrective code generation with text-based simulation. When given a task by the user, the code generator first produces the initial version of the robot operation code. Then our LLM-based simulator produces an observation of the robot's trajectory through text-based simulation of the code. After that, the evaluator analyzes the observation together with the task and generates feedback that depicts the mismatches (if any) in the robot's actions. Based on the feedback, the code generator regenerates the code that corrects the mismatches. This iterative ``generation-simulation-evaluation" loop continues until the evaluation confirms task objectives are achieved or the maximum number of iterations is reached. The following sections present the detailed design of our code generation, text-based simulation, and evaluation.

\subsection{Code Generation} \label{sec:code gen}
The code generation in our method is achieved by configuring an LLM agent (code generator) with a robot operation-related system prompt. We adopt the strategies in the GSCE framework~\cite{GSCE}, which enhance the reasoning capabilities of LLMs in generating UAV operation code that aligns with task instructions. Additionally, the code generator regenerates the code that resolves the mismatches according to the feedback from the evaluator in section~\ref{sec:evaluation}. In detail, given a task description or evaluation feedback, the code generator produces a robot operation code that aims to accomplish the task or resolve the mismatches.

\subsection{Static Text-Based Simulation using LLM} \label{sec:LLM simulation}
Given the robot operation code generated by the code generator, the goal of our static text-based simulation is to accurately interpret the robot's actions, reason about the state transitions, predict the robot states, and generate an observation of the robot's trajectory. Specifically, we formulate our text-based simulation using LLM as \(O = \langle S, A, T, C\rangle\), where \(O\) denotes the robot trajectory observation produced by the simulation, \(S = (s_0, s_1, s_2, \dots, s_n)\) denotes the finite set of discrete states, \(A\) denotes the finite set of robot actions, \(T: S \times A \rightarrow S\) denotes the state transition, and \(C\) denotes the text of robot operation code.  The trajectory observation \(O\) captures a sequence of robot actions \(l = (a_1, a_2, \dots, a_n)\) that transitions robot from the initial state \(s_0\) through intermediate states, to the final state \(s_n\). \(O\) also embeds the history of the robot's trajectory dynamics for each robot action \(l\) and their transitions \(T\) after executing code \(C\). This observation enables the subsequent evaluation to identify the mismatches and correct them in the next code generation. Therefore, the reliability of LLM-driven robotics depends critically on the accuracy of \(O\).


To generate \(O\), our design leverages the semantic reasoning capabilities of LLMs to simulate the execution of \(C\) by interpreting its textual content rather than executing the code in a real simulator. The LLM implements a function \(F: C \times S \times A \rightarrow S\) as a simulator that maps from a given code, current state, and action to the next state. Specifically, upon receiving a code script, the LLM-simulator simulates the execution of code, interprets the actions in the code, reasons about the corresponding state transitions, and predicts the next state. When the simulation completes, the LLM outputs its observation of the robot's trajectory as the simulation outcome.

To further enhance the reliability and accuracy of our text-based simulation, we further design a system prompt framework. The system prompt of LLM-simulator is composed of \textit{role}, \textit{APIs}, \textit{policies}, and \textit{examples} \footnote{The detailed design of the system prompt is provided in Appendix A}. 

\begin{itemize}
    \item \textbf{Role:} Defines the LLM agent responsible for simulating code execution by analyzing the provided code, inferring the intended actions, and outputting a description of the robot's trajectory. 
    \item \textbf{APIs:} Provide definitions of robot action that guide the LLM in understanding the code's intent, enabling inference actions \(l\) and their corresponding state transitions \(T\) encoded within the text of the code \(C\).
    \item \textbf{Policies:} Instruct the LLM with the code execution policies where the LLM lacks prior knowledge. The policies clarify the APIs usage, state transition rules, and environmental settings.
    \item \textbf{Examples:} Consist of pairs of code and trajectory that demonstrate how a robot operation code script \(C\) should be interpreted into semantic observation \(O\). These examples guide the LLM via few-shot learning, enabling the LLM-simulator to generate observations \(O\) that follow the same style and structure.
\end{itemize}

As demonstrated in Fig~\ref{fig:design}, the structured system prompt enables the LLM simulation to accurately produce an observation of the UAV actions, which facilitates the subsequent evaluation to identify the mismatches.

\subsection{Evaluation} \label{sec:evaluation}
In evaluation, the evaluator identifies the mismatches (if any) between the observation \(O\) and the task description, and then provides feedback specifying the mismatched actions. To ensure evaluation accuracy, we adopted the evaluator design in~\cite{CLGSCE}, which has demonstrated effectiveness in identifying the deviations between the UAV's trajectory and task description. The feedback from the evaluation provides the code generator with a clearer understanding of the objectives implied by the task description, thereby steering the generation of corrected code to better align with the task.

\section{EXPERIMENT} \label{sec:experiment}

\subsection{Experiment Setup} \label{sec:experiment setup}

\subsubsection{\textbf{Experimental Environment}}
We implement our proposed method and comparison method using OpenAI ``o3-mini'' (o3-mini-2025-01-31)~\cite{o3mini} and ``o4-mini'' (o4-mini-2025-04-16)~\cite{o4mini} as the foundational LLMs. To measure the performance of the methods, the experiment is conducted on a quadcopter on both the ``simple\_flight'' flight controller in AirSim \cite{AirSim} and the PX4 flight controller~\cite{PX4} in Gazebo~\cite{gazebo} \footnote{Simulator configurations and setups are provided in Appendix B}. During the experiment, UAV state information was accessible from the simulators and utilized for performance measurement purposes. Furthermore, all experiments are averaged over three repetitions to mitigate the randomness of LLM generation \cite{LLMrandomness}. 

\subsubsection{\textbf{Task Dataset}}
For the experiments, we adopt the Advanced task set from~\cite{CLGSCE} as the benchmark dataset to measure the performance. The Advanced task set contains 20 UAV operation tasks with varying levels of complexity, each involving 6-19 actions to reach the goal state. The tasks are designed to emulate real-world UAV operation scenarios that require complex reasoning about the UAV's state in the world environment, such as flying complex geometric patterns with scenario requirements.  To ensure the authenticity of the result, all tasks are manually validated in both AirSim and Gazebo to avoid potential simulation-induced errors that could affect the experiment result.

\subsubsection{\textbf{Compared Method}}

We compare our method against four methods:

\begin{itemize}
    \item \textbf{Direct Analysis (Direct):} Uses the semantic checker from~\cite{AutoTAMP}, where an LLM agent directly analyzes whether the generated code aligns with the task and provides feedback for code correction.
    \item \textbf{Simulator-based (Numerical):} Dynamically executes the generated code in a simulator to obtain numerical state observations, which are then evaluated to provide feedback for code correction~\cite{InteractivePlanning}.
    \item \textbf{Simulator-based (Semantic):} A SOTA method extends the Numerical method \cite{InteractivePlanning} by transforming numerical state observations produced from the simulator into semantic trajectory descriptions~\cite{CLGSCE} to improve the code correction efficiency. During the experiments, we modified the transformation algorithm to accommodate different robots and simulators. 
\end{itemize}

\subsection{Evaluation Setup}

\subsubsection{\textbf{Evaluation Metrics}}
Following~\cite{CLGSCE}, we evaluate performance using \textbf{Completeness} and \textbf{Success Rate (SR)}. 

\textbf{Completeness} measures the proportion of actions in a given task that are executed correctly. It is computed as the ratio between the number of correctly executed actions and the total number of actions in the ground-truth sequence. This metric provides insight into performance throughout the intermediate execution process. For a given task \(i\), the completeness is defined as:
\begin{equation}
    \mathrm{Completeness}_i = \frac{|a^{\mathrm{correct}}_i|}{|l^{\mathrm{gt}}_i|}
    \label{eq:completeness_i}
\end{equation}

where (\(a^{\mathrm{correct}}_i\)) denotes the count of actions correctly executed for task \(i\), and (\(l^{\mathrm{gt}}_i\)) is the total number of actions in the task's ground truth. The overall completeness is averaged over \(n\) tasks.

\textbf{SR} reflects task-level reliability by measuring whether the robot successfully reaches the final goal state while following the correct sequence of actions that produce the intended state transitions. A task is considered successful only if the complete trajectory is executed without errors (\(SR_i = 1, \text{ if } \mathrm{Completeness}_i \equiv 1\)).

\subsubsection{\textbf{Ground Truth}}
The ground truth is represented by a list of state transitions. Each state transition is a vector of four elements: \([x, y, z, \theta]\), where \(x\), \(y\), and \(z\) denote the robot's position changes in the North, East, and Down axes, and \(\theta\) represents yaw rotation.

\subsection{Result and Analysis}
\subsubsection{\textbf{Overall Result}}
The overall performance of our proposed method and comparison methods on both simple\_flight controller in AirSim and PX4 controller in Gazebo are summarized in Table~\ref{tab:arisim results} and Table~\ref{tab:px4 results}. For both LLM models, our LLM-simulator consistently supports reliable corrective code generation across different robot configurations, achieving success rates above $85\%$ on the simple\_flight controller and $86.7\%$ on the PX4 controller. These results demonstrate both the reliability of our text-based simulation and its adaptability across diverse robot systems.

In particular, our method achieves performance comparable to the SOTA Semantic method~\cite{CLGSCE} without requiring dynamic code execution in simulators that are explicitly designed to support robot simulations. This highlights that the proposed LLM-simulator can reliably conduct static text-based simulation of robot code to support corrective code generation. Furthermore, our method outperforms the Numerical method~\cite{InteractivePlanning}, indicating that the semantic observations generated by our LLM-simulator capture richer semantics about the robot's trajectory dynamics than the numerical representations. This enables an equally effective correction process as the SOTA Semantic method \cite{CLGSCE}, but without the need for customizing algorithms for different robot configurations to transform numerical states into semantic descriptions. In contrast, the Direct method~\cite{AutoTAMP} yields unreliable performance, proving the limitations of directly configuring LLMs to analyze code and highlighting the necessity to strengthen LLM's capabilities for LLM-based simulation.

\newcolumntype{L}[1]{>{\raggedright\arraybackslash}p{#1}}
\newcolumntype{C}[1]{>{\centering\arraybackslash}m{#1}}
\newcolumntype{R}[1]{>{\raggedleft\arraybackslash}p{#1}}

\begin{table}[!htbp]
    \centering
    \caption{Results of UAV with Simple\_Flight Controller}
        \begin{tabular}{lC{.8cm}cC{.8cm}c}
            \toprule
            & \multicolumn{2}{c}{o3-mini} & \multicolumn{2}{c}{o4-mini}\\
            \midrule
            & SR & Completeness & SR & Completeness\\
            \midrule
            Direct \cite{AutoTAMP} & $43.3\%$ & $70.9\%$  & $33.3\%$ & $57.6\%$ \\
            Numerical \cite{InteractivePlanning} & $73.3\%$ & $92.4\%$  & $81.7\%$ & $96.1\%$ \\
            Semantic \cite{CLGSCE} & $\mathbf{85.0}\%$ & $\mathbf{98.5}\%$  & $88.3\%$ & $98.1\%$ \\
            Ours       & $ \mathbf{85.0}\%$ & $ 97.0\%$ & $\mathbf{90.0\%}$ & $\mathbf{98.3}\%$ \\
            \bottomrule
        \end{tabular}
    \label{tab:arisim results}
\end{table}

\begin{table}[!htbp]
    \centering
    \caption{Results of UAV with PX4 Controller}
        \begin{tabular}{lC{.8cm}cC{.8cm}c}
            \toprule
            & \multicolumn{2}{c}{o3-mini} & \multicolumn{2}{c}{o4-mini}\\
            \midrule
            &  SR & Completeness & SR & Completeness\\
            \midrule
            Direct \cite{AutoTAMP} & $55.0\%$ & $77.5\%$ & $25.0\%$ & $50.9\%$ \\
            Numerical \cite{InteractivePlanning} & $83.3\%$ & $97.1\%$ & $86.7\%$ & $96.9\%$ \\
            Semantic \cite{CLGSCE} & $\mathbf{88.3}\%$ & $\mathbf{98.3}\%$ & $\mathbf{93.3}\%$ & $\mathbf{98.6}\%$ \\
            Ours       & $86.7\%$ & $97.7\%$ & $\mathbf{93.3}\%$ & $96.9\%$ \\
            \bottomrule
        \end{tabular}
    \label{tab:px4 results}
\end{table}

\subsubsection{\textbf{Text-Based Simulation Accuracy}} \label{sec:observation accuracy}
To further evaluate the reliability of our text-based simulation, we compare the accuracy of the trajectory observations generated by the LLM-simulator against those obtained using the Semantic method~\cite{CLGSCE} in the dynamic simulator. For each task in the Advanced task set, we construct a corresponding ground truth code ($C^{\text{correct}}$) that implements the task. The LLM-simulator is then used to statically simulate the execution of each $C^{\text{correct}}_i$ and produce trajectory observations. The accuracy of the simulation is computed as \(TP/N \times 100\%\), where $TP$ denotes the number of observations that accurately capture the robot trajectory dynamics, and $N$ is the total number of simulated code in $C^{\text{correct}}$. 

As shown in Table~\ref{tab:observation accuracy}, the LLM-simulator achieves $97.5\%$ accuracy on ``o3-mini'' and $100\%$ on ``o4-mini'' in capturing robot trajectory dynamics. These results demonstrate that the proposed LLM-simulator can reliably simulate the execution of robot actions, reason over the state transitions, and predict subsequent robot states. Therefore, the observation from the text-based simulation accurately reflects the intended robot's trajectory dynamics from the code, enabling the evaluator to effectively detect mismatched actions in the code. 



\begin{table}[!thbp] 
    \centering
    
    \caption{Text-Based Simulation Observation Accuracy}
    \begin{tabular}{l c c}
        \toprule
        & o3-mini & o4-mini\\
        \midrule
        Semantic \cite{CLGSCE} & $100.0\%$ & $100.0\%$ \\
        Ours & $97.5\%$ & $100.0\%$\\
        \bottomrule
    \end{tabular}
    \label{tab:observation accuracy}
\end{table}

\subsubsection{\textbf{Text-Based Simulation Evaluation Accuracy}} \label{sec:evaluation accuracy}

\begin{table}[!thbp] 
    \centering
    
    \caption{Evaluation Accuracy over Observations from LLM-Simulator}
    \begin{tabular}{l c c c c c}
        \toprule
        & \multicolumn{2}{c}{o3-mini} & \multicolumn{2}{c}{o4-mini} & Avg.\\
        \midrule
        & $C^{\text{correct}}$ & $C^{\text{incorrect}}$ & $C^{\text{correct}}$ & $C^{\text{incorrect}}$ \\
        \midrule
        Semantic \cite{CLGSCE} & $90.0\%$ & $91.7\%$ & $93.3\%$ & $93.3\%$ & $92.1\%$\\
        Ours & $91.7\%$ & $90.0\%$ & $93.3\%$ & $91.7\%$ & $91.7\%$\\
        \bottomrule
    \end{tabular}
    \label{tab:evaluation accuracy}
\end{table}

We further analyze whether the observations generated by the LLM-simulator can effectively support the evaluation process. Specifically, we compare the evaluation accuracy when using observations generated by our LLM-simulator against those produced from the Semantic method~\cite{CLGSCE}. This experiment leverages the ground-truth code set $C^{\text{correct}}$ in section~\ref{sec:observation accuracy}, along with an additional set $C^{\text{incorrect}}$ that contains at least one error action in code for each task. The evaluation accuracy is computed as \((TP + TN)/N \times 100\%\), where $TP$ denotes the number of correct evaluations on $C^{\text{correct}}$ and $C^{\text{incorrect}}$ (i.e., correctly identifying whether the trajectory matches with the task and providing a detailed explanation if mismatches are identified), and $N$ is the total number of evaluated code scripts. 

As presented in Table~\ref{tab:evaluation accuracy}, our method achieves over $90\%$ accuracy on ``o3-mini'' and $91.7\%$ accuracy on ``o4-mini''. The results are comparable to the Semantic method~\cite{CLGSCE} without the need for modifying algorithms for each robot and simulator to transform numerical states into semantic descriptions. The results also demonstrate that the observations from the LLM-simulator are sufficient to support effective evaluation as the SOTA Semantic method~\cite{CLGSCE} and are adaptable to different robot systems.

However, due to the non-determinism of LLM generation~\cite{Non-determinism}, identical actions may yield semantically equivalent but syntactically different observations. For example, the action ``fly 5 meters south'' from the text-based simulation may be produced as ``followed by 5 meters back (returning in the north-south direction)''. While such descriptions remain interpretable to humans, their implicit ambiguity can cause misinterpretations for LLMs~\cite{ambiguity-misinterpret}. Consequently, observations generated by the Semantic method~\cite{CLGSCE} exhibit slightly higher evaluation accuracy, as they are deterministic and free from linguistic variations. 

\subsection{Ablation Study on LLM-Simulator Design}

We conduct an ablation study on the design of our LLM-simulator. Specifically, we investigate how different components of the LLM's system prompt affect the overall system performance. As specified in Section~\ref{sec:evaluation}, the \textit{role} defines the LLM to be a simulator, and the \textit{APIs} provide definitions of robot action APIs, they serve as the essential component to ensure the LLM performs the text-based simulation. Thus, they are retained in the ablation study. For the remaining components, we remove \textit{policies}, \textit{examples}, and both \textit{policies} and \textit{examples} to measure their impact on the overall system performance. The results in Table~\ref{tab:ablation} show that removing either \textit{policies} or \textit{examples} leads to degradation in performance, while removing both yields the largest decline. Furthermore, removing \textit{examples} produces a larger negative impact than removing \textit{policies}, which is consistent with prior evidence that LLMs exhibit strong few-shot learning capabilities \cite{llm-fewshot}. The results of the ablation study highlight that both instructing the LLM with code execution rules (\textit{policies}) and providing demonstrations (\textit{examples}) are essential for enhancing the performance of the LLM-simulator, and their combination yields the best overall performance.

\begin{table}[!htbp]
    \centering
    \caption{Overall System Performance over LLM-Simulator Design}

        \begin{tabular}{lcccc}
            \toprule
            & \multirow{2}{*}{Ours} & w/o & w/o & w/o policies \\
            &  & policies & examples &  \& examples \\
            \specialrule{.8pt}{2pt}{2pt}
            \multicolumn{5}{c}{o3-mini}\\
            \midrule
            SR & $\mathbf{85.0\%}$ & $83.3\%$ & $81.7\%$ & $71.7\%$\\
            Completeness & $\mathbf{97.0\%}$ & $94.9\%$ & $94.0\%$ & $90.8\%$\\
            \specialrule{.8pt}{2pt}{2pt}
            \multicolumn{5}{c}{o4-mini}\\
            \midrule
            SR & $\mathbf{91.7\%}$ & $83.3\%$ & $83.3\%$ & $68.3\%$ \\
            Completeness & $\mathbf{97.7\%}$ & $96.1\%$ & $95.1\%$ & $93.1\%$ \\
            \bottomrule
        \end{tabular}

    \label{tab:ablation}
\end{table}

\section{REAL-WORLD DEPLOYMENT}
We validate our method through the deployment of physical robotic platforms. For consistency, the deployment is conducted using OpenAI's ``o3-mini'' and ``o4-mini'' models, and the performance of robot deployment is measured using the same metrics defined in section~\ref{sec:experiment setup}.

\subsection{Robot Setup}
\subsubsection{\textbf{UAV}}
We deploy our method on the Holybro X500 V2 quadcopter equipped with a Pixhawk 6X flight controller running PX4 firmware. For the task set, we select two tasks from each complexity level of the Advanced task set, resulting in a total of 8 tasks for deployment. In the deployment, the user types the task description into a ground station computer that runs our method to generate the UAV operation code (the computer is connected to the Internet for accessing the OpenAI API). The generated code is then transmitted to the flight controller via MAVSDK~\cite{mavsdk} for task execution. During the flight, the UAV's state information is collected through MAVSDK for performance measurement.

\subsubsection{\textbf{Ground Vehicle}}
We further evaluate the method on a ROSMASTER X3 ground vehicle controlled by ROS~\cite{ROS}. We then design 8 tasks for the ground vehicle to form patterns when driving on the ground. In the deployment, the user accesses the onboard computer remotely and types in the task descriptions, then the onboard computer (connected to the Internet for accessing the OpenAI API) runs our method to generate the ground vehicle operation code and then executes the code to control the vehicle's movement. 

\subsection{Result and Analysis}
The results of the real-world deployment are summarized in Table~\ref{tab:robot deployment}. Overall, our method demonstrates consistent performance across both UAV and ground vehicle platforms, validating the effectiveness of the proposed LLM-simulator in real-world settings. On both the UAV and ground vehicle tasks, our approach achieves high success rates and completeness, confirming the reliability and adaptability of our framework to different robot systems. These findings highlight that the static text-based simulation framework is reliable in supporting corrective code generation in real-world robot execution.

\begin{table}[!thbp] 
    \centering
    \caption{Robot Deployment Performance}
    \begin{tabular}{l c c c c c}
        \toprule
        & \multicolumn{2}{c}{o3-mini} & \multicolumn{2}{c}{o4-mini}\\
        \midrule
        & SR & Completeness & SR & Completeness\\
        \midrule
        UAV & $87.5\%$ & $98.6\%$ & $91.7\%$ & $97.9\%$ \\
        Ground Vehicle & $87.5\%$ & $96.9\%$ & $87.5\%$ & $97.2\%$ \\
        \bottomrule
    \end{tabular}
    \label{tab:robot deployment}
\end{table}


\section{CONCLUSIONS}
This paper presented an enhanced LLM-Driven corrective robot operation code generation framework. Different from existing solutions that require dynamic execution in a physical or simulation environment for code feedback and refinement, our framework is designed with a novel static text-based simulation solution powered by LLM, and hence addresses the challenges brought by the configuration of a dynamic code execution environment and potential long execution time for refinement.  
The experiment results on both different UAV systems validated the performance of our simulation solution in terms of both simulation accuracy and the reliability of robot operation code generation. Moreover, real-world deployments on physical robots further demonstrated the adaptability of our framework across different configurations and environments.

\section*{APPENDIX}

\subsection{LLM-Simulator System Prompt} \label{apdx: system prompt}

\fbox{\parbox{.95\linewidth}{
\textbf{Role:} 
You will analyze and infer the intention of the provided Python drone control code, then generate a description of the drone actions in one paragraph.

You should focus on the code, not the comments. Because code will be the actual actions of the drone.
}}

\fbox{\parbox{.95\linewidth}{
\textbf{APIs:}
Here are the available functions for the drone when you infer the drone actions and their state transitions:

\textit{aw.takeoff()} - takes off the drone.

\textit{aw.land()} - lands the drone.

\textit{aw.fly\_to([x, y, z])} - flies the drone to the position specified as a list of three arguments corresponding to world XYZ coordinates.

\textit{aw.get\_yaw()} - returns the current yaw of the drone in degrees.

\textit{aw.set\_yaw(yaw)} - sets the yaw of the drone to the specified value in degrees.

\textit{aw.get\_drone\_position()} - returns the current position of the drone as a list of 3 floats corresponding to world XYZ coordinates.
}}

\fbox{\parbox{.95\linewidth}{
\textbf{Policies:}
Important drone coordinate directional information and action policies:

1. The horizontal axes are Y and X, the vertical axis is Z. 

2. When rotating the drone, turning right or clockwise means positive, the yaw angle should increase.

3. \textit{aw.fly\_to([x, y, z])} function uses NED coordinate system (world coordinates), positive X axis is North/forward, positive Y axis is East/right, positive Z axis is Down. When flying up, the Z value should decrease. When flying down, the Z value should increase.

4. The drone is initialized facing north (Yaw = 0 degrees).

5. Map Yaw angle degree from -180 to 180, for example: map 270 to -90.
}}

\fbox{\parbox{.95\linewidth}{
\textbf{Examples:} 
Below are some examples; you should follow the output format in these examples in your answers.

Query: ``

\textit{current\_position = aw.get\_drone\_position()}

\textit{aw.fly\_to([current\_position[0], current\_position[1], current\_position[2] - 10])}
''

Answer:

``The drone flies 10 meters up."

Query: ``

\textit{aw.takeoff()}

\textit{current\_position = aw.get\_drone\_position() }

\textit{aw.fly\_to([current\_position[0], current\_position[1], current\_position[2] - 5])}

\textit{current\_position = aw.get\_drone\_position()}

\textit{aw.set\_yaw(90)}

\textit{aw.fly\_to([current\_position[0] + 5, current\_position[1], current\_position[2]])}

\textit{current\_position = aw.get\_drone\_position()}

\textit{current\_yaw = get\_yaw()}

\textit{aw.set\_yaw(90 + current\_yaw)}

\textit{aw.fly\_to([current\_position[0], current\_position[1] + 5, current\_position[2]])}

\textit{current\_position = aw.get\_drone\_position()}

\textit{current\_yaw = get\_yaw()}

\textit{aw.set\_yaw(90 + current\_yaw)}

\textit{aw.fly\_to([current\_position[0] - 5, current\_position[1], current\_position[2]])}

\textit{current\_position = aw.get\_drone\_position()}

\textit{current\_yaw = get\_yaw()}

\textit{aw.set\_yaw(90 + current\_yaw)}

\textit{aw.fly\_to([current\_position[0], current\_position[1] - 5, current\_position[2]])}
''

Answer:

``First, the drone takes off and climbs 5 m. Next, it rotates 90 degrees clockwise (now facing east) and flies to a point 5 m north along the world X axis. It then rotates 90 degrees clockwise (now facing south), and flies 5 m west along the world Y axis. Then rotate 90 degrees clockwise (now facing west), fly 5 m south along world X axis. Then rotate 90 degrees clockwise (now facing north) and fly 5 m east along world Y axis.''
}}

\subsection{Simulator Configurations} \label{apdx:simulator}
In our experiments, we employ two widely used simulators: AirSim~\cite{AirSim} and Gazebo~\cite{gazebo}.

\subsubsection{AirSim} Developed by Microsoft, AirSim provides a high-fidelity virtual environment for research in autonomous systems such as UAVs, self-driving cars, and ground robots. Built on Unreal Engine~\cite{unreal}, it supports realistic rendering of robot dynamics. In our experiments, we use the pre-built ``block''~\cite{Blocks} environment, which provides a clear and open airspace suitable for UAV testing.

\subsubsection{Gazebo} Gazebo~\cite{gazebo} models physical dynamics such as gravity, friction, and contact forces, making it applicable to both aerial and ground robots. It integrates seamlessly with the ROS operating system~\cite{ROS}. For our experiments, we use the PX4~\cite{PX4} software-in-the-loop (SITL) setup to control a simulated quadcopter within a virtual world environment of Gazebo.



\bibliographystyle{IEEEtran}
\bibliography{ref}

\end{document}